\documentclass[letterpaper]{article} 
\usepackage{aaai23}  
\usepackage{times}  
\usepackage{helvet}  
\usepackage{courier}  
\usepackage[hyphens]{url}  
\usepackage{graphicx} 
\usepackage{natbib}  
\usepackage{caption} 
\usepackage{algorithm}
\usepackage{algorithmic}
\usepackage{amsthm}
\usepackage{amsmath}
\usepackage{amsfonts}

\usepackage{microtype}
\usepackage{verbatim}
\usepackage{multirow}
\usepackage{threeparttable}
\usepackage{graphicx}
\usepackage{booktabs}
\usepackage{subfig}
\usepackage{comment}
\usepackage[title]{appendix}
\usepackage{adjustbox}
\usepackage{newfloat}
\usepackage{listings}
\DeclareCaptionStyle{ruled}{labelfont=normalfont,labelsep=colon,strut=off} 
\floatstyle{ruled}
\newfloat{listing}{tb}{lst}{}
\floatname{listing}{Listing}
\newcommand*\samethanks[1][\value{footnote}]{\footnotemark[#1]}
\title{Orders Are Unwanted: Dynamic Deep Graph Convolutional Network for Personality Detection}
\author{
    Tao Yang\textsuperscript{\rm 1\thanks{Equal contribution.}}, Jinghao Deng\textsuperscript{\rm 1\samethanks[1]}, Xiaojun Quan\textsuperscript{\rm 1}\thanks{Corresponding author.}, Qifan Wang\textsuperscript{\rm 2}
}
\affiliations{
    \textsuperscript{\rm 1} School of Computer Science and Engineering, Sun Yat-sen University \\
    \textsuperscript{\rm 2} Meta AI \\
    \{yangt225, dengjh27\}@mail2.sysu.edu.cn, quanxj3@mail.sysu.edu.cn, wqfcr@fb.com \\
}

\usepackage{bibentry}

\begin{document}

\maketitle

\begin{abstract}
Predicting personality traits based on online posts has emerged as an important task in many fields such as social network analysis. One of the challenges of this task is assembling information from various posts into an overall profile for each user. While many previous solutions simply concatenate the posts into a long document and then encode the document by sequential or hierarchical models, they introduce unwarranted orders for the posts, which may mislead the models. In this paper, we propose a dynamic deep graph convolutional network (D-DGCN) to overcome the above limitation. Specifically, we design a learn-to-connect approach that adopts a dynamic multi-hop structure instead of a deterministic structure, and combine it with a DGCN module to automatically learn the connections between posts. The modules of post encoder, learn-to-connect, and DGCN are jointly trained in an end-to-end manner. Experimental results on the Kaggle and Pandora datasets show the superior performance of D-DGCN to state-of-the-art baselines. Our code is available at \url{https://github.com/djz233/D-DGCN}.
\end{abstract}

\section{Introduction}
\label{intro}
Text-based personality detection is an emerging task in computational psycho-linguistics and affective computing \cite{jiang2019automatic}. The objective is to identify one's personality traits based on the texts she/he creates. Personality detection contributes meaningful cues to explaining an individual's behavior, emotion, and motivation \cite{mehta2019recent, zhang2019persemon}. The availability of a tremendous amount of social media posts containing users' digital traces provides rich data sources for this task and has aroused the interest of NLP and psychology researchers \cite{cui2017survey,xue2018deep,keh2019myers,amirhosseini2020machine,lynn2020hierarchical}.

\begin{figure}[t]
	\centering
    \vspace{0.4cm}
	\includegraphics[width=0.99\linewidth]{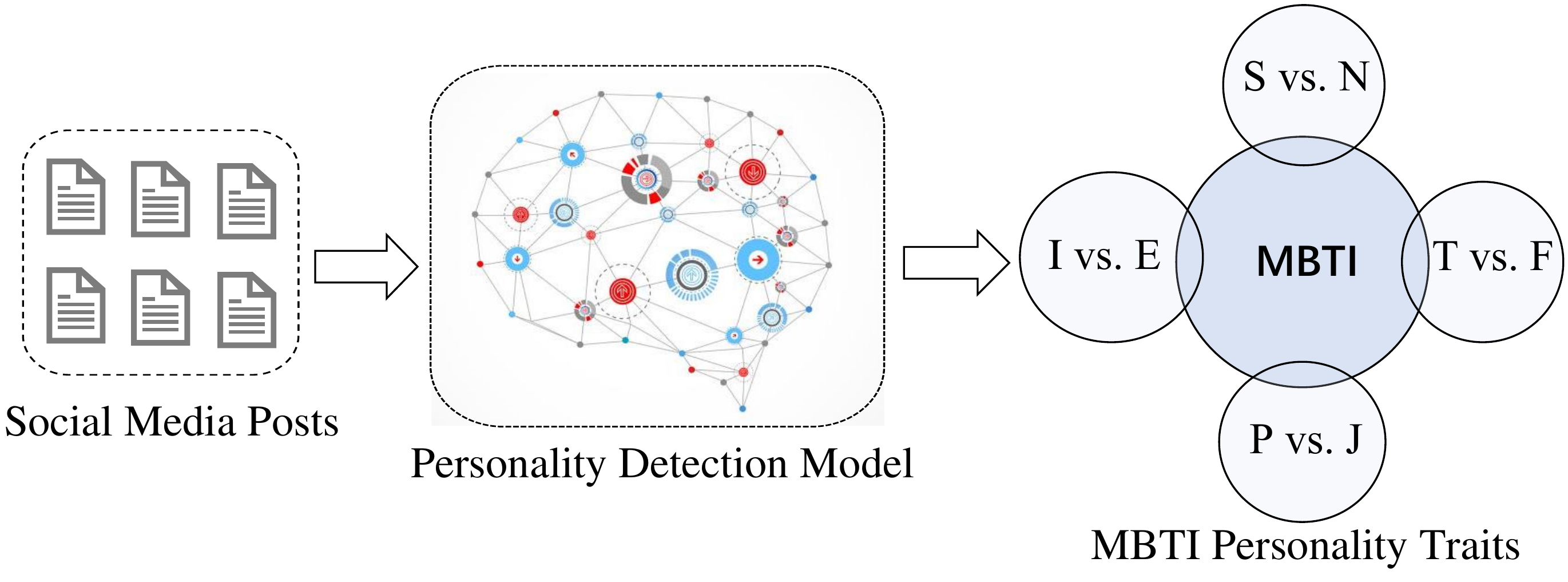}
	\caption{An example of personality detection for MBTI. The input are multiple social media posts from a single user and the output are her/his personality traits along dimensions of \textit{\textbf{I}ntroversion} vs. \emph{\textbf{E}xtroversion}, \textit{\textbf{S}ensing} vs. \emph{i\textbf{N}tuition}, \textit{\textbf{T}hink} vs. \emph{\textbf{F}eeling}, and \textit{\textbf{P}erception} vs. \emph{\textbf{J}udging}.}
	\label{figure1}
\end{figure}

The taxonomy of personality is generally defined along different dimensions. Figure \ref{figure1} shows an example of personality detection in the Myers-Briggs Type Indicator (MBTI) taxonomy \cite{myers1991introduction}. Although it can be naturally regarded as a multi-label classification task, personality detection has new challenges. First, the input of this task is not a single document but a set of posts, which are usually topic-agnostic short documents. Second, not every post necessarily contains personality clues, so the key is how to piece together useful information in different posts into a representative user profile. These posts can be arbitrarily merged into a long document and encoded sequentially \cite{jiang2019automatic,zhou2019gear}, or they can be encoded separately and then aggregated into a user representation by hierarchical networks \cite{lynn2020hierarchical,xue2018deep}. In either case, however, orders are introduced among the posts. Intuitively, these posts should work complementarily to build the user personality profile rather than treated sequentially or hierarchically. 

The graph is a natural structure to represent the posts in an unordered way, whereas it tends to be non-trivial to predefine the implicit personality-aware connections between posts as previous work of graph neural networks. In this paper, we propose a novel graph-based post fusion model, namely dynamic deep graph convolutional network (D-DGCN), to address the above issues. D-DGCN builds a graph to represent the posts of a user, in which each post is represented as a node and initialized by the embedding from a pre-trained language model such as BERT \cite{devlin2018bert}. A special node denoting the user is also added to facilitate personality classification. Moreover, considering that the connection between two nodes is not established deterministically, and inspired by GraphMask \cite{schlichtkrull2020interpreting}, we propose a novel multi-hop-based learn-to-connect (L2C) module with a differentiable threshold function to automatically build the essential edges layer by layer.~L2C enables the model to be more adaptive and scalable to different samples. Then, in light of the problem that conventional graph convolutional networks (GCNs) cannot stack too many layers because of the over-smoothing issue, a deep graph convolutional network (DGCN) \cite{liu2020towards} is applied to gather useful information from the posts from larger receptive fields. DGCN remedies over-smoothness by decoupling the transformation and propagation operations in GCNs, allowing the network to stack deeper.  
Extensive experiments are conducted on the Kaggle and Pandora datasets, and the results show that D-DGCN outperforms the baseline methods. Besides, extensive ablation studies and analysis further demonstrate that the L2C and DGCN modules play an indispensable role in the performance boosts.

The contributions are summarized as follows.

\begin{itemize}
	\item We propose a novel D-DGCN to tackle the post order issue, which is critical in personality detection.

	\item The proposed D-DGCN contains two interactive modules, L2C and DGCN, that interact intimately to learn how to establish connections between nodes layer by layer, enabling an unordered fusion of post information.
	
	\item We conduct extensive experiments on two benchmarks and demonstrate that the proposed D-DGCN outperforms all baselines and establishes a new state of the art.
\end{itemize}

\section{Related Work}\label{sec:work}
\subsection{Personality Detection}
Traditional work on personality detection relies heavily on feature engineerings \cite{yarkoni2010personality,schwartz2013personality}, such as extracting psycholinguistic features from Linguistic Inquiry and Word Count (LIWC) \cite{pennebaker2001linguistic}. These features are then fed into such machine-learning models as support vector machines \cite{cui2017survey} and XGBoost \cite{amirhosseini2020machine}. Recently, deep learning methods have dominated the research of personality detection and various post encoding methods have appeared. \citet{jiang2019automatic} simply concatenated all the posts from a single user into a document and fed it into a pre-trained language model. \citet{xue2018deep} and \citet{gjurkovic2020pandora} used CNN to aggregate the post representations. \citet{lynn2020hierarchical} and \citet{wang2021personality} adopted a hierarchical attention network to generate the user representation from posts. However, these approaches introduce unnecessary orders into the posts, which are likely to be captured by the models and affect their generalization ability. Although \citet{yang2021multi} tried to store posts in the memory of Transformer-XL \cite{dai2019transformer} to fix the order issue, the interactions between posts entirely rely on the self-attention mechanism \cite{vaswani2017attention}, which is prone to introduce irrelevant information.

\subsection{Graph Convolution Network}
Graph Convolution Network (GCN) \cite{kipf2016semi}, which is a special kind of graph neural networks (GNNs), learns node feature representation by iteratively aggregating features from neighbors in a convolutional way. Although it has achieved notable success in many applications, GCN faces the problem of over-smoothness, resulting in rapid performance drops as the layers stack. Encouraging efforts have been made to address this issue. \citet{wu2019simplifying} proposed a simple SGC model by reducing unnecessary complexity in GCN. \citet{chen2020measuring} designed two metrics to quantize the smoothness and over-smoothness in node representations. \citet{liu2020towards} provided a theoretical analysis of this problem and came up with a deep GNN, namely DAGNN, by decoupling the propagation and transformation processes. Inspired by their work, our model employs a similar strategy to overcome the over-smoothing problem.

\begin{figure*}[htbp]
	\centering
	\includegraphics[width=0.95\textwidth]{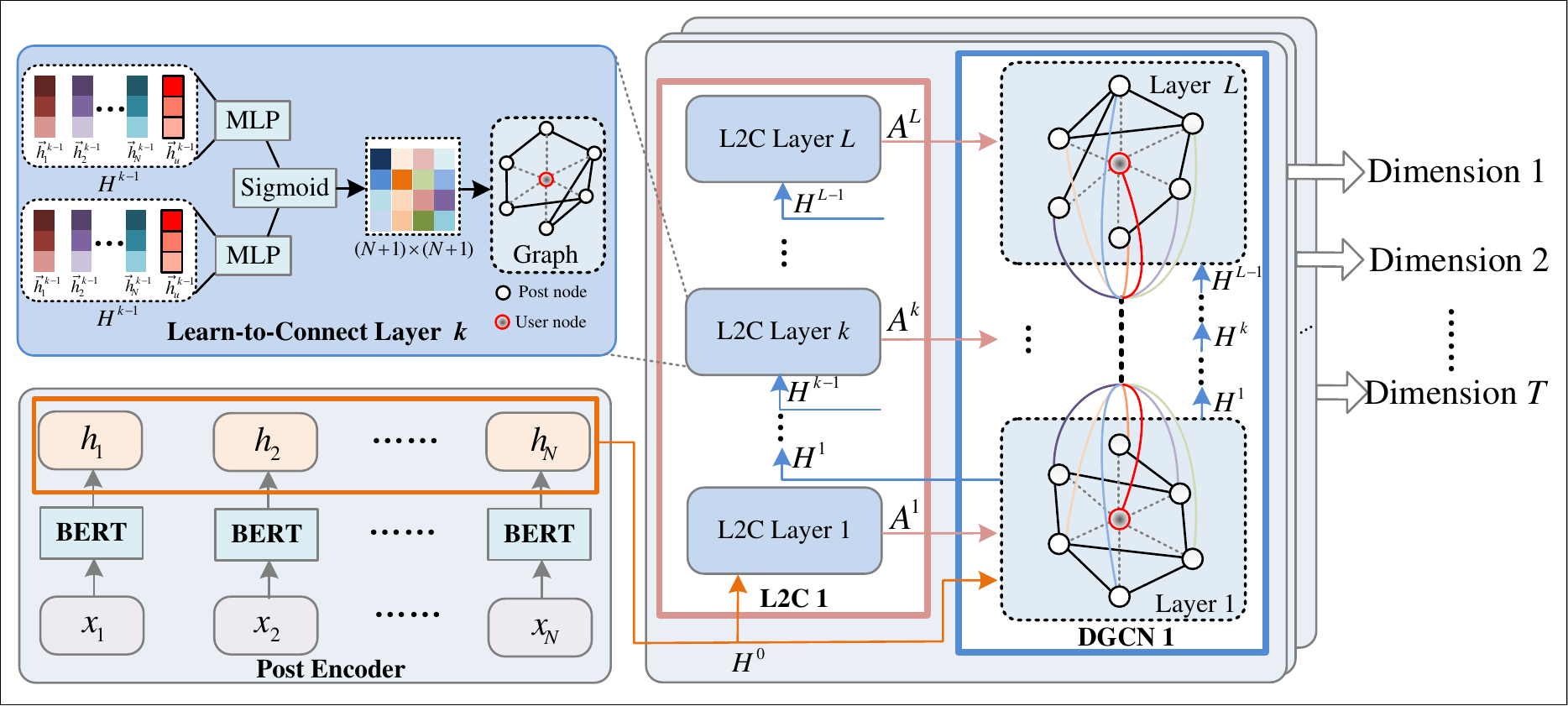}
	\caption{Architecture of our D-DGCN, which comprises a shared post encoder, $T$ L2C modules and $T$ DGCNs. The $T$ L2C modules and the $T$ DGCNs are parallelized to model $T$ personality dimensions.}
	\label{figure2}
\end{figure*}

\subsection{Graph Construction}
The graph can be constructed statically or dynamically in NLP tasks \cite{wu2021graph}. The static approach constructs the graph during preprocessing, which leverages linguistic knowledge and manually defined rules such as dependency parse trees \cite{wang2020relational}, knowledge graphs \cite{xie2020reinceptione, yang2021psycholinguistic}, and co-occurrence and document-word relations \cite{yao2019graph}. A graph constructed by the static approach is usually fixed and the structure cannot be optimized during graph representation learning, which could be sub-optimal. The dynamic approach learns the graph during training and the structure can be optimized end-to-end. Prior works construct dynamic graphs via similarity metrics \cite{chen2019graphflow,chen2020iterative} or attention mechanisms \cite{chen2019reinforcement}, and have shown the effectiveness of this approach in different tasks.

\section{Model Architecture}
We first formally define the personality detection task. Given a set $P = \left\{ {{p_1}, {p_2}, \ldots, {p_N}} \right\}$ of posts by a user, where ${p_i} = \left[ {{w_{i1}}, {w_{i2}}, \ldots, {w_{iM}}} \right]$ is the $i$-th post with $M$ tokens, the objective is to predict the personality traits $Y = \left\{ {{y^{(1)}}, {y^{(2)}}, ..., {y^{(T)}}} \right\}$ of this user along $T$ different dimensions. As the overall architecture shows in Figure \ref{figure2}, our D-DGCN mainly includes a pre-trained language model as post encoder and two interactive modules, namely learn-to-connect (L2C) and deep graph convolutional network (DGCN). The prediction of each personality dimension can be regarded as an individual classification task. Except for the shared post encoder, each classification task has respective L2C and DGCN modules so as to model trait-specific features. In the following, we take one of the classification tasks as an example to introduce D-DGCN.
\subsection{Post Encoder}
We employ BERT \cite{devlin2018bert} to encode each post separately. Note that applying BERT directly to personality detection with out-of-domain data may harm its performance. Previous work finds that incorporating domain knowledge into BERT is helpful to address the domain adaptation challenge \cite{gururangan2020don}. Therefore, we first post-train BERT via masked language model (MLM) on the training sets of Kaggle and Pandora. 

Formally, the context vector ${\vec h_i}$ for the $i$-th post $p_i$ is obtained by:
\begin{equation}
	{\vec h_i} = {\mathop{\rm BERT}\nolimits} \left( {{p_i}} \right) \in {\mathbb{R}^{1 \times d}}
\end{equation}
where ${\mathop{\rm BERT}\nolimits} \left(\cdot  \right)$ denotes the final hidden state of ``[CLS]" in post-trained BERT, and $d$ is the dimension of the output. As a result, we obtain a set of contextual representations $\hat{H} = \lbrace {{\vec h_1},{\vec h_2}, \cdots ,{\vec h_N}} \rbrace$ for the given $N$ posts of a user.

\subsection{Learn to Connect} \label{sec:l2c}
After getting all the post representations $\hat{H}$ for a user, we then embark on how to fuse them into a user profile representation. Unlike previous studies that generally fuse them sequentially or hierarchically, we employ a more plausible method, namely deep graph convolutional network (DGCN), to fuse them in an unordered way. Specifically, we first represent each user by a graph and each of his posts by a node in the graph. To generate a user representation facilitating personality classification, we put a special user node $u$ in the graph to aggregate information from other nodes. Therefore, the initial representations of all the nodes become $H = \lbrace \vec h_1,\vec h_2, \cdots ,\vec h_N, \vec h_u \rbrace \in \mathbb{R}^{(N+1)\times d}$, where $\vec h_u$ is the special node representation initialized by averaging the node representations in $\hat{H}$. 

Next, we introduce how to construct the edges of the graph in a learning approach. Since it is unclear how to define the connections between posts in favor of personality detection, inspired by GraphMask \cite{schlichtkrull2020interpreting}, we adopt a learning approach to establish the connections based on proper node representations. Specifically, we propose a dynamic graph learning module L2C with $L$ layers. As shown in Figure \ref{figure2}, each layer of L2C tries to adjust the learned graph dynamically based on the output of the previous layer of DGCN, and then passes the new graph back to DGCN to update node representations for the next layer.

Formally, the adjacency matrix $A^k$ used to represent the graph in the $k$-th layer is calculated as:
\begin{equation}
	A^k = {\rm{L2C}}\left( {H^{k - 1}} \right)\in {\mathbb{R}^{(N+1)\times (N+1)}}
\end{equation}
where $H^{k - 1}$ denotes the node representations produced by the $(k-1)$-th layer of DGCN, and ${\mathop{\rm L2C}\nolimits} \left(\cdot  \right)$ is the function to determine whether there is an edge between two nodes. As shown in Figure \ref{figure2}, two MLPs are introduced to implement this function. Letting $H^{k - 1}$ represent the query and key matrices, the L2C module computes an adjacency weight, $r_{ij}^k$, between nodes $i$ and $j$ as:
\begin{equation} \label{eq3}
	r_{ij}^k = \sigma \left( { Relu{\left( { \vec h_i^{k - 1} W_Q^k}  \right) } \left( \vec h_j^{k - 1} W_K^k \right)^{\rm{T}}} \right)
\end{equation}
where $\sigma$ is the sigmoid function, $\vec h_i^{k - 1}$ and $\vec h_j^{k - 1}$ are the representations of nodes $i$ and $j$ in the $(k-1)$-th layer of DGCN, $W_Q^k \in {\mathbb{R}^{d\times hid}}$  and $W_K^k \in {\mathbb{R}^{d\times hid}}$ are layer-specific linear transformations, and $d$ and $hid$ are the dimensions of post representation and hidden state of MLP, respectively. 

After obtaining the adjacency weight, a differentiable threshold function is utilized to determine whether there is an edge between two nodes, giving rise to the adjacency matrix $A^k$ for this layer. We implement the differentiable threshold function via a programming trick. Specifically, the implementation details can be summarized as follows. First, the adjacency weight $r_{ij}^k$ is normalized by a scaling factor that is almost equal to its own value and has no gradient:
\begin{equation} \label{trick}
	{\hat a_{ij}^k} = \frac{{r_{ij}^k}}{{\hat r_{ij}^k + \varepsilon }}
\end{equation}
where ${\hat r_{ij}^k}$ means detaching the gradient from ${r_{ij}^k}$ during the training stage and ${\varepsilon=1e-6}$ is used to prevent overflow. Then, we use a mask matrix to produce the final adjacency matrix. The element $a_{ij}^k$ in $A^k$ is calculated by:
\begin{equation} \label{eq5}
	a_{ij}^k = \left\{ {\begin{array}{*{20}{c}}
			{\hat a_{ij}^k}&{r_{ij}^k > {\mu}  }\\
			0&{{\rm{otherwise}}}
	\end{array}} \right.
\end{equation}	
where $\mu$ is a threshold and is set to 0.5 naturally. To better help the L2C module remove unnecessary edges, we introduce $\ell_{0}$ norm with Hard Concrete distribution \cite{louizos2017learning} to minimize the number of graph connections:
\begin{equation}
    \ell_{0} = \sum\limits_{k=1}^L \sum\limits_{(i,j) \in A^k} {a_{ij}^k}.
\end{equation}

Note that this implementation of the differentiable threshold function can regularize the gradients to some extent so that L2C can be fine-tuned layer by layer instead of being adjusted arbitrarily. In this way, L2C can automatically learn to establish connections between nodes effectively. The adjacency matrix $A^k$ is then fed into DGCN to generate the node representations for the $(k+1)$-th layer.  

\subsection{Deep Graph Convolutional Network}
After computing the adjacency matrix dynamically, post embeddings are fed into GCNs for encoding. As mentioned above, over-smoothness may limit the performance of our L2C module. To address this, we apply DGCN \cite{liu2020towards}, which alleviates the issue by decoupling the transformation and propagation operations in Eq.~(\ref{DGCN-enc}), allowing the L2C module to stack deeper so that the learned graph can be fine-tuned better, and reducing the parameters of GCNs to prevent overfitting. Specifically, GCNs perform propagation and transformation as follows:
\begin{equation} \label{gcn}
	{H^{k + 1}} = f\left( {\hat A{H^k}{W^{k}}} \right)
\end{equation}
where $\hat A = {D^{ - \frac{1}{2}}}(A+I){D^{ - \frac{1}{2}}}$ is the normalized symmetric adjacency matrix. For DGCN, it updates the node representations as follows:
\begin{equation} \label{DGCN-enc}
	{H^{k + 1}} = {\hat A^k{H^k}}.
\end{equation}

Compared to the original formula of GCNs in Eq.~(\ref{gcn}), Eq.~(\ref{DGCN-enc}) doesn't have learnable transformation matrix $W^{k}$ and contains only the propagation operation. After $L$ layers of iteration, we obtain the node representations for every layer:
\begin{equation}
	{\bf{H}} = \left[ {{H^0},{H^1}, \cdots ,{H^{L}}} \right] \in {\mathbb{R}^{(N+1)\times (L+1) \times d}}
\end{equation}
where $H^0 = H$ are the initial node representations. Intuitively, $\bf{H}$ contains node information from both low and high layers. Then, a trainable projection vector $\vec c \in {\mathbb{R}^{d\times 1}}$ is employed to determine which layers are more useful:
\begin{equation}
	S = \sigma \left( {{\bf{H}} \cdot \vec c} \right)\in {\mathbb{R}^{(N+1)\times (L+1)\times 1}}
\end{equation}
\begin{equation}
	\tilde S = {\rm{Reshape}}\left( S \right) \in {\mathbb{R}^{(N+1)\times 1\times (L+1)}}
\end{equation}
where $\sigma$ is the sigmoid function, and ${\rm{Reshape}}\left(  \cdot  \right)$ is used to reshape a matrix for further computation. The eventual node representations $H^{out}$ are obtained by:
\begin{equation} \label{prop}
	{H^{out}} = {\tilde S \odot \bf{H}} \in {\mathbb{R}^{(N+1)\times d}}
\end{equation}
where $\odot$ is the multiplication of matrices of the last two dimensions. 


\subsection{Objective Function}
As mentioned in Learn to Connect, the output $H^{out}$ of DGCN contains a special node representation $\vec h_u^{out}$ for a user. Based on $\vec h_u^{out}$, we employ a linear transformation followed by a softmax function to predict each personality trait:
\begin{equation}
	y = {\mathop{\rm softmax}\nolimits} (\vec h_u^{out}{W_u} + {b_u})
\end{equation}
\begin{equation} \label{ce_loss}
	\ell_{ce} = \frac{1}{V}\sum\limits_{i = 1}^V {\sum\limits_{j = 1}^T {\left[ { - y_i^{j}\log p\left( {y_i^{j}|\theta } \right)} \right]} } 
\end{equation}
where $W_u \in {\mathbb{R}^{d \times 2}}$ is a trainable weight matrix, $b_u$ is a bias term, $V$ is the number of training samples, $y_i^{j}$ is the true label of the $j$-th personality dimension, and $p( {y_i^{j}|\theta })$ is the predicted probability for this dimension under parameters $\theta$. We use the cross-entropy loss function for all the $T$ personality traits. The whole objective function is defined as:
\begin{equation} \label{total_loss}
    \ell_{total} = \lambda \ell_{ce} +  \sum\limits_{i = 1}^V {\sum\limits_{j = 1}^T \ell_{0}} 
\end{equation}
where $\lambda$ is an adaptive Lagrange multiplier. In this way, L2C and DGCN can be jointly optimized. 

\section{Experiments}
In this section, we introduce the settings of our experiments and report the overall results.

\subsection{Datasets} 
Following previous studies \cite{gjurkovic2020pandora, yang2021multi, yang2021learning, yang2021psycholinguistic}, we choose the Kaggle\footnote{\url{https://www.kaggle.com/datasnaek/mbti-type}} and Pandora\footnote{\url{https://psy.takelab.fer.hr/datasets/all}} MBTI datasets for our evaluations. While the former is collected from PersonalityCafe\footnote{\url{http://personalitycafe.com/forum}}, with 45-50 social media posts included for each of 8675 users, the latter is collected from Reddit\footnote{\url{https://www.reddit.com}}, with dozens to hundreds of social media posts for each of 9067 users. As previous work \cite{yang2021multi, yang2021learning, yang2021psycholinguistic}, we remove words that match any personality type from the posts. Since the two datasets are severely imbalanced, we employ the \emph{Macro-F1} metric to measure the performance. Table \ref{tab:dataset} shows the distribution and split of personality and amount of used posts in the two datasets.

\begin{table}[t]
    \begin{adjustbox}{width=1.0\columnwidth,center}
        \begin{tabular}{c|cccc}
            \toprule
            Dataset & Types   & Train         & Validation   & \multicolumn{1}{c}{Test} \\ 
            \midrule
            \multirow{5}{*}{Kaggle}  & I / E & 4011 / 1194 & 1326 / 409 & 1339 / 396              \\
            & S / N & 610 / 4478  & 222 / 1513 & 248 / 1487\\
            & T / F & 2410 / 2795 & 791 / 944  & 780 / 955 \\
            & P / J & 3096 / 2109 & 1063 / 672 & 1082 / 653 \\ 
            & Posts & 246794 & 82642 & 82152 \\ \midrule
            \multirow{5}{*}{Pandora} & I / E & 4278 / 1162 & 1427 / 386 & 1437 / 377    \\
            & S / N & 727 / 4830  & 208 / 1605 & 210 / 1604 \\
            & T / F & 3549 / 1891  & 1120 / 693 & 1182 / 632 \\
            & P / J & 3211 / 2229 & 1043 / 770 & 1056 / 758  \\ 
            & Posts & 523534 & 173005 & 174080 \\ 
            \bottomrule
        \end{tabular}
        \end{adjustbox}
         \caption{Statistics of the Kaggle and Pandora datasets.}
        \label{tab:dataset}
\end{table}

\begin{table*}[t]
	\label{table2}
	    \setlength{\tabcolsep}{0.15cm}
	\renewcommand{\arraystretch}{1.1}
    \begin{adjustbox}{width=0.9\linewidth,center}
	\begin{tabular}{l|ccccc|ccccc}
		\toprule
		\multirow{2}{*}{\textbf{Methods}}& \multicolumn{5}{c|}{\textbf{Kaggle}}&\multicolumn{5}{c}{\textbf{Pandora}}\\ 
		\cline{2-6}
		\cline{6-11}
		& \textit{\textbf{I} / \textbf{E}} & \textit{\textbf{S} / \textbf{N}} &  \textit{\textbf{T} / \textbf{F}} & \textit{\textbf{P} / \textbf{J}} &  \textbf{Avg} & \textit{\textbf{I} / \textbf{E}} & \textit{\textbf{S} / \textbf{N}} &  \textit{\textbf{T} / \textbf{F}} & \textit{\textbf{P} / \textbf{J}} & \textbf{Avg} \\
		\hline
		SVM & 53.34 & 47.75 & 76.72 & 63.03 & 60.21 & 44.74 & 46.92 & 65.37 & 56.32 & 53.34\\
		{XGBoost} & 56.67 & 52.85 & 75.42 & 65.94 & 62.72 & 45.99 & 48.93 & 66.38 & 55.55 & 54.21\\
		\hline
		$\rm{{LSTM_{mean}}}$ & 57.82 & 57.87 & 69.97 & 57.01 & 60.67 & 48.01 & 52.01 & 63.48 & 56.12 & 54.91 \\
		SN+Attn & 62.34 & 57.08 & 69.26 & 63.09 & 62.94 & 54.60 & 49.19 & 61.82 & 53.64 & 54.81 \\
		\hline
		$\rm{{BERT_{concat}}}$ & 58.33 & 53.88 & 69.36 & 60.88 & 60.61 & 54.22 & 49.15 & 58.31 & 53.14 & 53.71\\
		$\rm{{BERT_{LSTM}}}$ & 58.12 & 51.44 & 70.02 & 55.92 & 58.88 & 52.70 & 47.92 & 62.27 & 49.97 & 53.22 \\
		$\rm{{BERT_{CNN}}}$ & 58.17 & 53.87 & 75.66 & 54.05 & 60.44 & 50.08 & 51.34 & 61.72 & 51.33 & 53.62 \\
		$\rm{{BERT_{mean}}}$ & 63.50 & 55.34 & 78.55 & 66.06 & 65.86 & 53.35 & 50.56 & 64.06 & 56.83 & 56.20\\
		$\rm{{BERT_{att}}}$ & 63.76 & 58.32 & 77.99 & 65.42 & 66.37 & 56.03 & 53.81 & 67.47 & 58.57 & 58.97 \\
		\hline
		TrigNet$^{*}$ & \textbf{69.54} & 67.17 & 79.06 & 67.69 & 70.86 & 56.69 & 55.57 & 66.38 & 57.27 & 58.98 \\
		Transformer-MD$^{*}$ & 66.08 & \textbf{69.10} & 79.19 & 67.50 & 70.47 & 55.26 & \textbf{58.77} & 69.26 & \textbf{60.90} & 61.05 \\
		D-DGCN & 68.41 & 65.66 & 79.56 & 67.22 & 70.21(70.33) & \textbf{61.55} & 55.46 & \textbf{71.07} & 59.96 & \textbf{62.01(62.49)}\\
		D-DGCN+$\ell_{0}$ & 69.52 & 67.19 & \textbf{80.53} & \textbf{68.16} & \textbf{71.35(71.59)} & 59.98 & 55.52 & 70.53 & 59.56 & 61.40(61.50)\\		
		\bottomrule
	\end{tabular}
    \end{adjustbox}
	\caption{Overall results of our D-DGCN and baseline models in {Macro-F1} (\%) score. * means the results are cited from original papers, and both reported the highest results. We report the mean score of D-DGCN after running with three random seeds. We also report the highest average-F1 scores in the bracket for making fair comparisons with existing SOTA. Best results are highlighted in bold.}
	\label{tab:overall}
\end{table*}

\subsection{Baselines} 
To make comprehensive evaluations and comparisons, we adopt the following models as our baselines:

\begin{itemize}
	\item \textbf{SVM} \cite{cui2017survey} and \textbf{XGBoost} \cite{amirhosseini2020machine}: The posts of a user are firstly concatenated into a document. Then, a SVM or XGBoost is employed for personality classification based on features extracted using bag-of-words methods.
	\item $\rm{\mathbf{LSTM_{mean}}}$ \cite{cui2017survey} and $\rm{\mathbf{BERT_{mean}}}$ \cite{keh2019myers}: BiLSTM or BERT is firstly utilized to encode each post, and the averaged post representation is used to represent each user.
	\item $\rm{\mathbf{BERT_{concat}}}$ \cite{jiang2019automatic}: This method simply concatenates the posts of a user into a long document and feeds it into BERT.
	\item $\rm{\mathbf{{BERT_{CNN}}}}$ \cite{gjurkovic2020pandora} and $\rm{\mathbf{{BERT_{LSTM}}}}$: The two methods are similar to $\rm{{BERT_{mean}}}$ but utilize CNN or LSTM to fuse the encoded posts.
	\item $\rm{\mathbf{{BERT_{att}}}}$: This method is similar to $\rm{{BERT_{mean}}}$, except that the post representations are summarized by attention instead of the mean pooling strategy. 
	\item \textbf{SN+Attn} \cite{lynn2020hierarchical}: This method employs a hierarchical attention network with both word-level and post-level attentions.
	\item \textbf{Transformer-MD} \cite{yang2021multi}: Transformer-MD encodes the posts sequentially by BERT and stores them in memory, which allows posts to access the information of former ones.
	\item \textbf{TrigNet} \cite{yang2021psycholinguistic}: TrigNet applies a modified GAT to fuse the posts and constructs the graph with psycholinguistic knowledge in LIWC.
\end{itemize}

\subsection{Implementation Details}
\label{sec:appendix_A}
All the deep learning models are implemented in PyTorch and trained with Adam \cite{kingma2014adam}. We set the maximum length of a post to 70 for both datasets and the maximum number of posts to 50 for Kaggle and 100 for Pandora. For non-pretrained baselines, we use 300-dimensional GloVe word embeddings \cite{pennington2014glove}. For pre-trained models, BERT is first initialized by BERT-BASE-CASED \cite{devlin2018bert} and then post-trained via standard MLM on two million posts selected from the training sets of Kaggle and Pandora with learning rate 6e-5 and batch size 256.

For the L2C module, we set the learning rate of MLPs to 1e-5. We set the learning rate of $\lambda$ in Eq.~(\ref{total_loss}) to 1e-2, and initialize $\lambda$ to 5.0 and limit its value between 0.0 to 100.0. After tuning on the validation dataset, we set the learning rate of BERT to 1e-5 and the dropout to 0.1, while the learning rate and dropout of other components are set to 1e-3 and 0.2, respectively. The hidden size of BiLSTM is set to 300. The depth $L$ of DGCN firstly changes from 1 to 6 with step size 1 and then changes from 6 to 24 with step size 3. We train D-DGCN for 25 epochs with 3 random seeds.

\subsection{Overall Results}
Table \ref{tab:overall} presents the overall results, in which the baselines are organized into four groups: shallow models (SVM and XGBoost), non-pretrained model ($\rm{{LSTM_{mean}}}$ and SN+Attn), pre-trained models with simple methods ($\rm{{BERT_{concat}}}$, $\rm{{BERT_{CNN}}}$, $\rm{{BERT_{LSTM}}}$, $\rm{{BERT_{mean}}}$, and $\rm{{BERT_{att}}}$) and with complex methods (Transformer-MD and TrigNet). We also present two types of our model D-DGCN and D-DGCN+$\ell_{0}$, whose objective functions are Eq.~(\ref{ce_loss}) and Eq.~(\ref{total_loss}), respectively. Our observations from the table are as follows. First, our models achieve the highest average F1 scores, verifying the effectiveness of our D-DGCN. Specifically, in the pre-trained setting, compared with traditional $\rm{{BERT_{att}}}$ and SN+Attn, D-DGCN achieves significant boosts in average F1 on the Kaggle and Pandora datasets. The boosted performance comes from two aspects: strong DGCN without introducing unnecessary order and a well-defined graph learning method. Compare with the two complex methods, D-DGCN and D-DGCN+$\ell_{0}$ also get better performance and show their strong competitiveness. Second, the result of D-DGCN family models shows the difference between the two datasets. D-DGCN+$\ell_{0}$ gets better performance in Kaggle, revealing that posts of Kaggle contain more noise.~While D-DGCN performs better on Pandora, it seems that L2C should preserve more information about Pandora.

Surprisingly, the shallow models SVM and XGBoost achieve comparable performance with certain deep learning models such as $\rm{{LSTM_{mean}}}$, and even $\rm{{BERT_{concat}}}$ and $\rm{{BERT_{CNN}}}$, showing that deep models are not omnipotent in personality detection and should be combined with appropriate encoding methods. Furthermore, among the five pre-trained models with simple methods, $\rm{{BERT_{att}}}$ and $\rm{{BERT_{mean}}}$ don't introduce order information and achieve higher scores, verifying our proposition.
\begin{table*}[t]
	\label{table3}
	\renewcommand{\arraystretch}{1.25}
    \begin{adjustbox}{width=0.85\linewidth,center}
	\begin{tabular}{l|ccccc|ccccc}
		\toprule
		\multirow{2}{*}{\textbf{Methods}}&
		\multicolumn{5}{c|}{\textbf{Kaggle}}&\multicolumn{5}{c}{\textbf{Pandora}}\\ 
		\cline{2-6}
		\cline{6-11}
		& \textit{\textbf{I} / \textbf{E}} & \textit{\textbf{S} / \textbf{N}} &  \textit{\textbf{T} / \textbf{F}} & \textit{\textbf{P} / \textbf{J}} &  \textbf{Avg} & \textit{\textbf{I} / \textbf{E}} & \textit{\textbf{S} / \textbf{N}} &  \textit{\textbf{T} / \textbf{F}} & \textit{\textbf{P} / \textbf{J}} & \textbf{Avg} \\
		\hline
		D-DGCN/MTGNN & 67.43 & 62.98 & 78.10 & \textbf{67.96} & 69.12 & 58.94 & 47.36 & 68.06 & 59.05 & 58.35 \\
		D-GCN & 64.62 & 62.23 & 79.01 & 64.77 & 67.66 & 55.67 & 55.00 & 69.19 & 57.51 & 59.34\\
		D-GAT & 66.28 & 63.74 & \textbf{80.93} & 67.54 & 69.62 & 60.34 & 53.12 & 68.18 & 59.43 & 60.27 \\
		\hline
		DGCN/fix & 66.49 & 62.60 & 78.86 & 61.90 & 67.46 & 58.87 & 53.71 & 63.76 & 53.90 & 57.56 \\
		D-DGCN/single-hop & 67.58 & 60.08 & 81.53 & 67.09 & 69.07 & 59.87 & 55.66 & 69.36 & 57.81 & 60.67\\
		D-DGCN/undirected & \textbf{69.17} & 59.76 & 78.15 & 67.68 & 68.69 & 60.41 & \textbf{56.26} & 69.60 & 58.39 & 61.16\\
		D-DGCN/-u & 66.85 & 64.14 & 79.14 & 64.76 & 68.72 & 58.28 & 54.51 & 68.83 & 56.44 & 59.51\\
		D-DGCN/-DAPT & 67.37 & 64.47 & 80.51 & 66.10 & 69.61 & 58.28 & 55.88 & 68.50 & 57.72 & 60.10\\
		\hline
		D-DGCN & 67.22 & \textbf{65.81} & 80.57 & 66.81 & \textbf{70.11} & \textbf{62.31} & 55.45 & \textbf{70.32} & \textbf{59.51} & \textbf{61.90}\\
		\bottomrule
	\end{tabular}
    \end{adjustbox}
	\caption{Results of ablation studies in {Macro-F1} (\%) score, where different fusion methods and different settings of D-DGCN are implemented for comparison.}
	\label{table:ablation}
\end{table*}

\section{Analysis}
In this section, we conduct extensive evaluations and provide thorough analysis and discussions. 

\subsection{Impact of Order}
\begin{table}[t]
	\renewcommand{\arraystretch}{1.2}
	\centering
	\resizebox{0.4\textwidth}{!}{
	\begin{tabular}{l|cc}
		\toprule
		
		\textbf{Methods}& \makebox[0.12\textwidth][c]{\textbf{Kaggle}} &\makebox[0.12\textwidth][c]{\textbf{Pandora}}\\ 
		\hline
		$\rm{{BERT_{concat}}}$ & 60.61 & 53.71 \\
		$\rm{{BERT_{concat/rd}}}$ & \textbf{61.30} & \textbf{54.92}  \\
            \hline
		$\rm{{BERT_{LSTM}}}$ & 58.88 & 53.22 \\
		$\rm{{BERT_{LSTM/rd}}}$  & \textbf{59.41} & \textbf{53.34} \\
		\bottomrule
	\end{tabular}
	}
	\caption{Results of different post orders in averaged {Macro-F1} (\%) score. Merging posts with random orders (/rd) performs better than initial orders.}	
	\label{table4}
	\vspace{-0.3cm}
\end{table}

To investigate the impact of post order, we conduct experiments with two representative order-sensitive models, namely $\rm{{BERT_{concat}}}$ (sequential) and $\rm{{BERT_{LSTM}}}$ (hierarchical). We randomly disrupt the initial post orders (by publication time) and re-train the two models ($\rm{{BERT_{concat/rd}}}$ and $\rm{{BERT_{LSTM/rd}}}$) for five times. The results in Table \ref{table4} show that $\rm{{BERT_{concat/rd}}}$ and $\rm{{BERT_{LSTM/rd}}}$ do not become worse but perform better than $\rm{{BERT_{concat}}}$ and $\rm{{BERT_{LSTM}}}$, suggesting that the initial post orders are non-essential information for personality detection.

\begin{figure*}[t]
	\centering
	\includegraphics[width=1\textwidth]{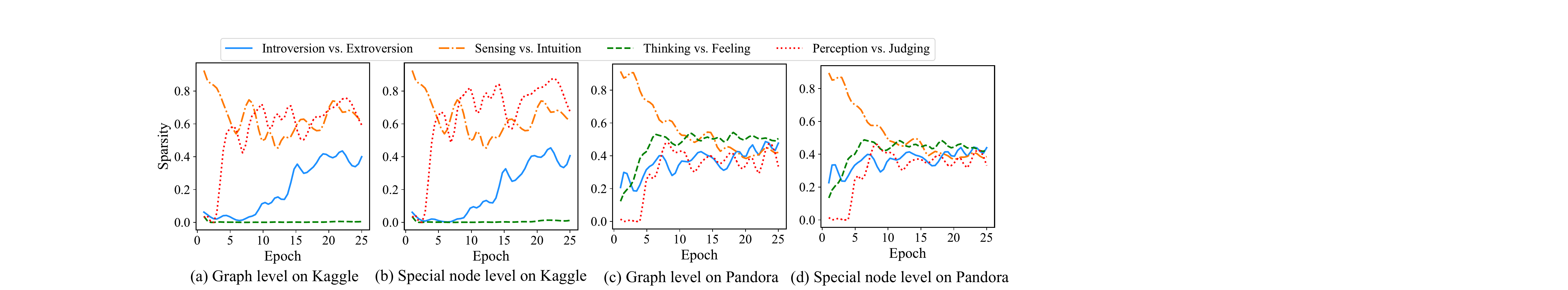}
	\caption{Curves of edge sparsity at the graph level and the special node ($u$) level on Kaggle and Pandora. The sparsity is calculated by the ratio of valid edges to total edges that can produce connections. Here, we set the layer of D-DGCN to 1 for the sake of simplicity.}
	\label{fig:sparse}
	\vspace{-0.2cm}
\end{figure*}

\subsection{Ablation Studies}
\subsubsection{Module Comparison}
Our D-DGCN comprises two main modules: L2C and DGCN, which represent the graph learning method and graph neural network.~To demonstrate our model’s superiority, we replace the two modules respectively.~We replace L2C module with MTGNN \cite{wu2020connecting} (D-DGCN/MTGNN) first.~From Table \ref{table:ablation}, it is clear that the new model D-DGCN/MTGNN underperforms D-DGCN by a considerable margin, showing the power of L2C in modeling implicit personality connections. To make a comparison among different graph neural networks, we replace DGCN with GCN (D-GCN) and GAT (D-GAT) in order. As the result shows, D-GCN suffers from over-smoothing issues and gets poorer performance. GAT \cite{velivckovic2018graph} does not have over-smoothing issues and is able to adjust the graph connection dynamically, which is an excellent substitute for GCN. However, D-GAT is still inferior to D-DGCN.

\subsubsection{Graph Construction Strategies}
The proposed L2C module relies on self-attention with a multi-hop structure and a differentiable threshold function to construct a directed and non-weighted graph dynamically. To investigate the effect of these factors, we make extensive comparisons from the following aspects:

\paragraph{Pre-defined Graph vs. Learned Graph}
The main difference between many previous graph-based models and our D-DGCN is that the former requires us to build a graph in advance. To verify the effectiveness, we implement DGCN/fix on pre-defined and fixed graphs constructed based on cosine similarities between posts that are encoded by Sentence-BERT \cite{reimers2019sentence}. We also implement a degenerating model of D-DGCN, D-DGCN/single-hop, which only learns the graph in the first layer. The two variants both share the graph in all layers. As shown in Table \ref{table:ablation}, DGCN/fix underperforms D-DGCN/single-hop apparently, implying that high-quality personality connections between posts are difficult to pre-define. Moreover, D-DGCN performs better than D-DGCN/single-hop, indicating that a single-hop structure is not enough to learn an effective graph for personality detection.

\paragraph{Undirected Graph vs. Directed Graph}
The graph learned by our L2C is directed since the connections established by Eq.~(\ref{eq3}) is asymmetric. To explore the necessity of a directed graph, we implement the undirected one, D-DGCN/undirected, by constraining the learned adjacency matrix to symmetry. As the result shows, D-DGCN/undirected shows poorer performance than the directed one (D-DGCN). The reason is probably that the directed graph contains richer personality information than the undirected graph.

\paragraph{Pooling vs. Special Node}
~Our D-DGCN inserts a special node $u$ in the graph to aggregate information and directly uses its final representation for classification. We experimentally compare it with the conventional pooling approach (D-DGCN/-u) which averages all the node representations in the last layer. As shown in Table \ref{table:ablation}, D-DGCN with the special user node outperforms D-DGCN/-u, confirming the effectiveness of inserting the special user node.

\subsubsection{Domain Adaptivity}
Recall that we use the training sets of Kaggle and Pandora for domain adaptive post-training of our D-DGCN model. To verify the effectiveness and necessity of this post-training, we conduct an experiment by replacing the post encoder with the original BERT (D-DGCN/-DAPT). As expected, the results in Table \ref{table:ablation} show that the performance declines to some extent on the two datasets. This implies that the original BERT does not fully learn transferable personality knowledge during pre-training, making it worthwhile to collect more unannotated social media texts and design personality-related pre-training tasks in future work.

\subsection{Sparsity Analysis}
\label{sec:sparsity}
To investigate the amount of required information and learning difficulty of different personality dimensions and datasets, we visualize the sparsity curves in terms of different personality dimensions during the training process. For each dataset, two set of sparsity curves, i.e., the percentage of valid edges at the graph level and the percentage of valid edges at the special node ($u$) level, are plotted in Figure \ref{fig:sparse}. From Figure \ref{fig:sparse} (a) and (c), we note that the curves at the graph level converge to different percentages, illustrating that different personality dimensions need different amounts of information. In addition, Figure \ref{fig:sparse} (b) and (d) show that the curves at the special node level are close to the graph ones (Figure  \ref{fig:sparse} (a) and (c)), demonstrating the consistency between the special node and the whole graph. Finally, Figure \ref{fig:sparse} (a) and (b) show that the dimension (\emph{T} vs. \emph{F}) remains sparse during training on Kaggle, and it gets the highest F1 score among the four dimensions in Table \ref{tab:overall}, implying that D-DGCN can easily distinguish whether a person is thinking or feeling.

\section{Conclusion}
In this paper, we presented an unordered post fusion model, D-DGCN, for personality detection. It uses a deep graph convolutional network to try to piece together information from multiple posts into an overall user profile. Unlike previous work that generally requires deterministic graphs to be pre-defined, D-DGCN employs a learn-to-connect approach that learns to build the graphs. Experimental results on two datasets show D-DGCN outperforms the baseline models. Moreover, we conducted extensive ablation studies and analysis to verify the effectiveness of the L2C and DGCN modules. To sum up, proper connection of posts is essential information in personality detection, but not order. Finally, developing personality-related pre-training tasks is a promising direction for future work.

\section*{Acknowledgments}
This work was supported by the National Natural Science Foundation of China (No. 62176270), the Guangdong Basic and Applied Basic Research Foundation (No. 2023A1515012832), the Program for Guangdong Introducing Innovative and Entrepreneurial Teams (No. 2017ZT07X355), and the Foundation of Key Laboratory of Machine Intelligence and Advanced Computing of the Ministry of Education.

\bibliography{anthology,custom}

\begin{thebibliography}{40}
\providecommand{\natexlab}[1]{#1}

\bibitem[{Amirhosseini and Kazemian(2020)}]{amirhosseini2020machine}
Amirhosseini, M.~H.; and Kazemian, H. 2020.
\newblock Machine Learning Approach to Personality Type Prediction Based on the
  Myers--Briggs Type Indicator{\textregistered}.
\newblock \emph{Multimodal Technologies and Interaction}, 4(1): 9.

\bibitem[{Chen et~al.(2020)Chen, Lin, Li, Li, Zhou, and
  Sun}]{chen2020measuring}
Chen, D.; Lin, Y.; Li, W.; Li, P.; Zhou, J.; and Sun, X. 2020.
\newblock Measuring and Relieving the Over-Smoothing Problem for Graph Neural
  Networks from the Topological View.
\newblock In \emph{AAAI}, 3438--3445.

\bibitem[{Chen, Wu, and Zaki(2020)}]{chen2020iterative}
Chen, Y.; Wu, L.; and Zaki, M. 2020.
\newblock Iterative deep graph learning for graph neural networks: Better and
  robust node embeddings.
\newblock \emph{Advances in Neural Information Processing Systems}, 33:
  19314--19326.

\bibitem[{Chen, Wu, and Zaki(2019{\natexlab{a}})}]{chen2019graphflow}
Chen, Y.; Wu, L.; and Zaki, M.~J. 2019{\natexlab{a}}.
\newblock Graphflow: Exploiting conversation flow with graph neural networks
  for conversational machine comprehension.
\newblock \emph{arXiv preprint arXiv:1908.00059}.

\bibitem[{Chen, Wu, and Zaki(2019{\natexlab{b}})}]{chen2019reinforcement}
Chen, Y.; Wu, L.; and Zaki, M.~J. 2019{\natexlab{b}}.
\newblock Reinforcement learning based graph-to-sequence model for natural
  question generation.
\newblock \emph{arXiv preprint arXiv:1908.04942}.

\bibitem[{Cui and Qi(2017)}]{cui2017survey}
Cui, B.; and Qi, C. 2017.
\newblock Survey analysis of machine learning methods for natural language
  processing for MBTI Personality Type Prediction.
\newblock \emph{Final Report Stanford University}.

\bibitem[{Dai et~al.(2019)Dai, Yang, Yang, Carbonell, Le, and
  Salakhutdinov}]{dai2019transformer}
Dai, Z.; Yang, Z.; Yang, Y.; Carbonell, J.~G.; Le, Q.; and Salakhutdinov, R.
  2019.
\newblock Transformer-XL: Attentive Language Models beyond a Fixed-Length
  Context.
\newblock In \emph{Proceedings of the 57th Annual Meeting of the Association
  for Computational Linguistics}, 2978--2988.

\bibitem[{Devlin et~al.(2018)Devlin, Chang, Lee, and
  Toutanova}]{devlin2018bert}
Devlin, J.; Chang, M.-W.; Lee, K.; and Toutanova, K. 2018.
\newblock Bert: Pre-training of deep bidirectional transformers for language
  understanding.
\newblock \emph{arXiv preprint arXiv:1810.04805}.

\bibitem[{Gjurkovi{\'c} et~al.(2020)Gjurkovi{\'c}, Karan, Vukojevi{\'c},
  Bo{\v{s}}njak, and {\v{S}}najder}]{gjurkovic2020pandora}
Gjurkovi{\'c}, M.; Karan, M.; Vukojevi{\'c}, I.; Bo{\v{s}}njak, M.; and
  {\v{S}}najder, J. 2020.
\newblock PANDORA Talks: Personality and Demographics on Reddit.
\newblock \emph{arXiv preprint arXiv:2004.04460}.

\bibitem[{Gururangan et~al.(2020)Gururangan, Marasovi{\'c}, Swayamdipta, Lo,
  Beltagy, Downey, and Smith}]{gururangan2020don}
Gururangan, S.; Marasovi{\'c}, A.; Swayamdipta, S.; Lo, K.; Beltagy, I.;
  Downey, D.; and Smith, N.~A. 2020.
\newblock Don't stop pretraining: adapt language models to domains and tasks.
\newblock \emph{arXiv preprint arXiv:2004.10964}.

\bibitem[{Jiang, Zhang, and Choi(2019)}]{jiang2019automatic}
Jiang, H.; Zhang, X.; and Choi, J.~D. 2019.
\newblock Automatic Text-based Personality Recognition on Monologues and
  Multiparty Dialogues Using Attentive Networks and Contextual Embeddings.
\newblock \emph{arXiv preprint arXiv:1911.09304}.

\bibitem[{Keh, Cheng et~al.(2019)}]{keh2019myers}
Keh, S.~S.; Cheng, I.; et~al. 2019.
\newblock Myers-Briggs Personality Classification and Personality-Specific
  Language Generation Using Pre-trained Language Models.
\newblock \emph{arXiv preprint arXiv:1907.06333}.

\bibitem[{Kingma and Ba(2014)}]{kingma2014adam}
Kingma, D.~P.; and Ba, J. 2014.
\newblock Adam: A method for stochastic optimization.
\newblock \emph{arXiv preprint arXiv:1412.6980}.

\bibitem[{Kipf and Welling(2016)}]{kipf2016semi}
Kipf, T.~N.; and Welling, M. 2016.
\newblock Semi-supervised classification with graph convolutional networks.
\newblock \emph{arXiv preprint arXiv:1609.02907}.

\bibitem[{Liu, Gao, and Ji(2020)}]{liu2020towards}
Liu, M.; Gao, H.; and Ji, S. 2020.
\newblock Towards Deeper Graph Neural Networks.
\newblock In \emph{Proceedings of the 26th ACM SIGKDD International Conference
  on Knowledge Discovery \& Data Mining}, 338--348.

\bibitem[{Louizos, Welling, and Kingma(2017)}]{louizos2017learning}
Louizos, C.; Welling, M.; and Kingma, D.~P. 2017.
\newblock Learning sparse neural networks through $ L_0 $ regularization.
\newblock \emph{arXiv preprint arXiv:1712.01312}.

\bibitem[{Lynn, Balasubramanian, and Schwartz(2020)}]{lynn2020hierarchical}
Lynn, V.; Balasubramanian, N.; and Schwartz, H.~A. 2020.
\newblock Hierarchical Modeling for User Personality Prediction: The Role of
  Message-Level Attention.
\newblock In \emph{Proceedings of the 58th Annual Meeting of the Association
  for Computational Linguistics}, 5306--5316.

\bibitem[{Mehta et~al.(2019)Mehta, Majumder, Gelbukh, and
  Cambria}]{mehta2019recent}
Mehta, Y.; Majumder, N.; Gelbukh, A.; and Cambria, E. 2019.
\newblock Recent trends in deep learning based personality detection.
\newblock \emph{Artificial Intelligence Review}, 1--27.

\bibitem[{Myers-Briggs(1991)}]{myers1991introduction}
Myers-Briggs, I. 1991.
\newblock Introduction to Type: A Description of the Theory and Applications of
  the Myers-Briggs Indicator.
\newblock \emph{Consulting Psychologists: Palo Alto}.

\bibitem[{Pennebaker, Francis, and Booth(2001)}]{pennebaker2001linguistic}
Pennebaker, J.~W.; Francis, M.~E.; and Booth, R.~J. 2001.
\newblock Linguistic inquiry and word count: LIWC 2001.
\newblock \emph{Mahway: Lawrence Erlbaum Associates}, 71(2001): 2001.

\bibitem[{Pennington, Socher, and Manning(2014)}]{pennington2014glove}
Pennington, J.; Socher, R.; and Manning, C.~D. 2014.
\newblock Glove: Global vectors for word representation.
\newblock In \emph{Proceedings of the 2014 Conference on Empirical Methods in
  Natural Language Processing (EMNLP)}, 1532--1543.

\bibitem[{Reimers and Gurevych(2019)}]{reimers2019sentence}
Reimers, N.; and Gurevych, I. 2019.
\newblock Sentence-BERT: Sentence Embeddings using Siamese BERT-Networks.
  EMNLP-IJCNLP 2019-2019 Conference on Empirical Methods in Natural Language
  Processing and 9th International Joint Conference on Natural Language
  Processing.
\newblock In \emph{Proceedings of the Conference}, 3982--3992.

\bibitem[{Schlichtkrull, De~Cao, and
  Titov(2020)}]{schlichtkrull2020interpreting}
Schlichtkrull, M.~S.; De~Cao, N.; and Titov, I. 2020.
\newblock Interpreting graph neural networks for nlp with differentiable edge
  masking.
\newblock \emph{arXiv preprint arXiv:2010.00577}.

\bibitem[{Schwartz et~al.(2013)Schwartz, Eichstaedt, Kern, Dziurzynski,
  Ramones, Agrawal, Shah, Kosinski, Stillwell, Seligman
  et~al.}]{schwartz2013personality}
Schwartz, H.~A.; Eichstaedt, J.~C.; Kern, M.~L.; Dziurzynski, L.; Ramones,
  S.~M.; Agrawal, M.; Shah, A.; Kosinski, M.; Stillwell, D.; Seligman, M.~E.;
  et~al. 2013.
\newblock Personality, gender, and age in the language of social media: The
  open-vocabulary approach.
\newblock \emph{PlOS One}, 8(9): e73791.

\bibitem[{Vaswani et~al.(2017)Vaswani, Shazeer, Parmar, Uszkoreit, Jones,
  Gomez, Kaiser, and Polosukhin}]{vaswani2017attention}
Vaswani, A.; Shazeer, N.; Parmar, N.; Uszkoreit, J.; Jones, L.; Gomez, A.~N.;
  Kaiser, {\L}.; and Polosukhin, I. 2017.
\newblock Attention is all you need.
\newblock In \emph{Advances in Neural Information Processing Systems},
  5998--6008.

\bibitem[{Veli{\v{c}}kovi{\'c} et~al.(2018)Veli{\v{c}}kovi{\'c}, Cucurull,
  Casanova, Romero, Li{\`o}, and Bengio}]{velivckovic2018graph}
Veli{\v{c}}kovi{\'c}, P.; Cucurull, G.; Casanova, A.; Romero, A.; Li{\`o}, P.;
  and Bengio, Y. 2018.
\newblock Graph Attention Networks.
\newblock In \emph{International Conference on Learning Representations}.

\bibitem[{Wang et~al.(2020)Wang, Shen, Yang, Quan, and
  Wang}]{wang2020relational}
Wang, K.; Shen, W.; Yang, Y.; Quan, X.; and Wang, R. 2020.
\newblock Relational Graph Attention Network for Aspect-based Sentiment
  Analysis.
\newblock \emph{arXiv preprint arXiv:2004.12362}.

\bibitem[{Wang et~al.(2021)Wang, Sui, Zheng, Shi, and
  Cao}]{wang2021personality}
Wang, X.; Sui, Y.; Zheng, K.; Shi, Y.; and Cao, S. 2021.
\newblock Personality Classification of Social Users Based on Feature Fusion.
\newblock \emph{Sensors}, 21(20): 6758.

\bibitem[{Wu et~al.(2019)Wu, Souza, Zhang, Fifty, Yu, and
  Weinberger}]{wu2019simplifying}
Wu, F.; Souza, A.; Zhang, T.; Fifty, C.; Yu, T.; and Weinberger, K. 2019.
\newblock Simplifying Graph Convolutional Networks.
\newblock In \emph{International Conference on Machine Learning}, 6861--6871.

\bibitem[{Wu et~al.(2021)Wu, Chen, Shen, Guo, Gao, Li, Pei, and
  Long}]{wu2021graph}
Wu, L.; Chen, Y.; Shen, K.; Guo, X.; Gao, H.; Li, S.; Pei, J.; and Long, B.
  2021.
\newblock Graph neural networks for natural language processing: A survey.
\newblock \emph{arXiv preprint arXiv:2106.06090}.

\bibitem[{Wu et~al.(2020)Wu, Pan, Long, Jiang, Chang, and
  Zhang}]{wu2020connecting}
Wu, Z.; Pan, S.; Long, G.; Jiang, J.; Chang, X.; and Zhang, C. 2020.
\newblock Connecting the Dots: Multivariate Time Series Forecasting with Graph
  Neural Networks.
\newblock \emph{arXiv preprint arXiv:2005.11650}.

\bibitem[{Xie et~al.(2020)Xie, Zhou, Liu, and Huang}]{xie2020reinceptione}
Xie, Z.; Zhou, G.; Liu, J.; and Huang, X. 2020.
\newblock ReInceptionE: relation-aware inception network with joint
  local-global structural information for knowledge graph embedding.
\newblock In \emph{Proceedings of the 58th Annual Meeting of the Association
  for Computational Linguistics}, 5929--5939.

\bibitem[{Xue et~al.(2018)Xue, Wu, Hong, Guo, Gao, Wu, Zhong, and
  Sun}]{xue2018deep}
Xue, D.; Wu, L.; Hong, Z.; Guo, S.; Gao, L.; Wu, Z.; Zhong, X.; and Sun, J.
  2018.
\newblock Deep learning-based personality recognition from text posts of online
  social networks.
\newblock \emph{Applied Intelligence}, 48(11): 4232--4246.

\bibitem[{Yang et~al.(2021{\natexlab{a}})Yang, Quan, Yang, and
  Yu}]{yang2021multi}
Yang, F.; Quan, X.; Yang, Y.; and Yu, J. 2021{\natexlab{a}}.
\newblock Multi-document transformer for personality detection.
\newblock In \emph{Proceedings of the AAAI Conference on Artificial
  Intelligence}, volume~35, 14221--14229.

\bibitem[{Yang et~al.(2021{\natexlab{b}})Yang, Yang, Quan, and
  Su}]{yang2021learning}
Yang, F.; Yang, T.; Quan, X.; and Su, Q. 2021{\natexlab{b}}.
\newblock Learning to answer psychological questionnaire for personality
  detection.
\newblock In \emph{Findings of the Association for Computational Linguistics:
  EMNLP 2021}, 1131--1142.

\bibitem[{Yang et~al.(2021{\natexlab{c}})Yang, Yang, Ouyang, and
  Quan}]{yang2021psycholinguistic}
Yang, T.; Yang, F.; Ouyang, H.; and Quan, X. 2021{\natexlab{c}}.
\newblock Psycholinguistic Tripartite Graph Network for Personality Detection.
\newblock \emph{arXiv preprint arXiv:2106.04963}.

\bibitem[{Yao, Mao, and Luo(2019)}]{yao2019graph}
Yao, L.; Mao, C.; and Luo, Y. 2019.
\newblock Graph convolutional networks for text classification.
\newblock In \emph{Proceedings of the AAAI Conference on Artificial
  Intelligence}, volume~33, 7370--7377.

\bibitem[{Yarkoni(2010)}]{yarkoni2010personality}
Yarkoni, T. 2010.
\newblock Personality in 100,000 words: A large-scale analysis of personality
  and word use among bloggers.
\newblock \emph{Journal of Research in Personality}, 44(3): 363--373.

\bibitem[{Zhang, Peng, and Winkler(2019)}]{zhang2019persemon}
Zhang, L.; Peng, S.; and Winkler, S. 2019.
\newblock PersEmoN: a deep network for joint analysis of apparent personality,
  emotion and their relationship.
\newblock \emph{IEEE Transactions on Affective Computing}.

\bibitem[{Zhou et~al.(2019)Zhou, Han, Yang, Liu, Wang, Li, and
  Sun}]{zhou2019gear}
Zhou, J.; Han, X.; Yang, C.; Liu, Z.; Wang, L.; Li, C.; and Sun, M. 2019.
\newblock GEAR: Graph-based evidence aggregating and reasoning for fact
  verification.
\newblock \emph{arXiv preprint arXiv:1908.01843}.

\end{thebibliography}

\end{document}